\documentclass[10pt,twocolumn,letterpaper]{article}

\usepackage{wacv}              % To produce the CAMERA-READY version
\usepackage{graphicx}
\usepackage{amsmath}
\usepackage{amssymb}
\usepackage{booktabs}
\usepackage{multirow}

\usepackage[pagebackref,breaklinks,colorlinks]{hyperref}

\usepackage[capitalize]{cleveref}
\crefname{section}{Sec.}{Secs.}
\Crefname{section}{Section}{Sections}
\Crefname{table}{Table}{Tables}
\crefname{table}{Tab.}{Tabs.}

\def\wacvPaperID{2553} % *** Enter the WACV Paper ID here
\def\confName{WACV}
\def\confYear{2024}

\begin{document}

\title{Triplet Attention Transformer for Spatiotemporal Predictive Learning}

\author{Xuesong Nie$^\ddagger$ \quad
Xi Chen$^\dagger$ \quad
Haoyuan Jin$^\ddagger$ \quad
Zhihang Zhu$^\ddagger$ \quad
Yunfeng Yan$^{\ddagger\diamond}$ \quad
Donglian Qi$^\ddagger$ \\ \quad
$^\ddagger$Zhejiang University\\
$^\dagger$The University of Hong Kong\\
}
\maketitle

\footnote{$^\diamond$Corresponding author.}

%%%%%%%%% ABSTRACT
\begin{abstract}
Spatiotemporal predictive learning offers a self-supervised learning paradigm that enables models to learn both spatial and temporal patterns by predicting future sequences based on historical sequences. Mainstream methods are dominated by recurrent units, yet they are limited by their lack of parallelization and often underperform in real-world scenarios. To improve prediction quality while maintaining computational efficiency, we propose an innovative triplet attention transformer designed to capture both inter-frame dynamics and intra-frame static features. Specifically, the model incorporates the Triplet Attention Module (TAM), which replaces traditional recurrent units by exploring self-attention mechanisms in temporal, spatial, and channel dimensions. In this configuration: (i) temporal tokens contain abstract representations of inter-frame,  facilitating the capture of inherent temporal dependencies; (ii) spatial and channel attention combine to refine the intra-frame representation by performing fine-grained interactions across spatial and channel dimensions. Alternating temporal, spatial, and channel-level attention allows our approach to learn more complex short- and long-range spatiotemporal dependencies. Extensive experiments demonstrate performance surpassing existing recurrent-based and recurrent-free methods, achieving state-of-the-art under multi-scenario examination including moving object trajectory prediction, traffic flow prediction, driving scene prediction, and human motion capture.
\end{abstract}

%%%%%%%%% BODY TEXT
\section{Introduction}
\label{sec:intro}
Predicting the future is an innate ability possessed by humans, making it a challenging task for machines due to the complex inner laws of the chaotic world. Spatiotemporal predictive learning as a data-driven approach generates future sequences based on historical sequences, with extensive applications including weather forecasting~\cite{reichstein2019deep,reichstein2019deep}, human motion forecasting~\cite{babaeizadeh2017stochastic,wang2018rgb}, traffic flow prediction~\cite{fang2019gstnet,wang2019memory}, representation learning~\cite{lotter2016deep,jenni2020video}, and vision-based predictive control~\cite{finn2016unsupervised,gupta2022maskvit}. In contrast to supervised models that require annotated data, spatiotemporal predictive models can uncover complex spatial and temporal correlations in a self-supervised manner using massive unlabeled data. Spatiotemporal data as the most accessible resource, these methodologies offer potential as unsupervised pre-training paradigms for universal representation learning~\cite{castrejon2019improved,oprea2020review,tan2023temporal,tan2023openstl}. 

\begin{figure}[t]
\begin{center}
\scalebox{0.5}{\includegraphics[width=0.98\textwidth]{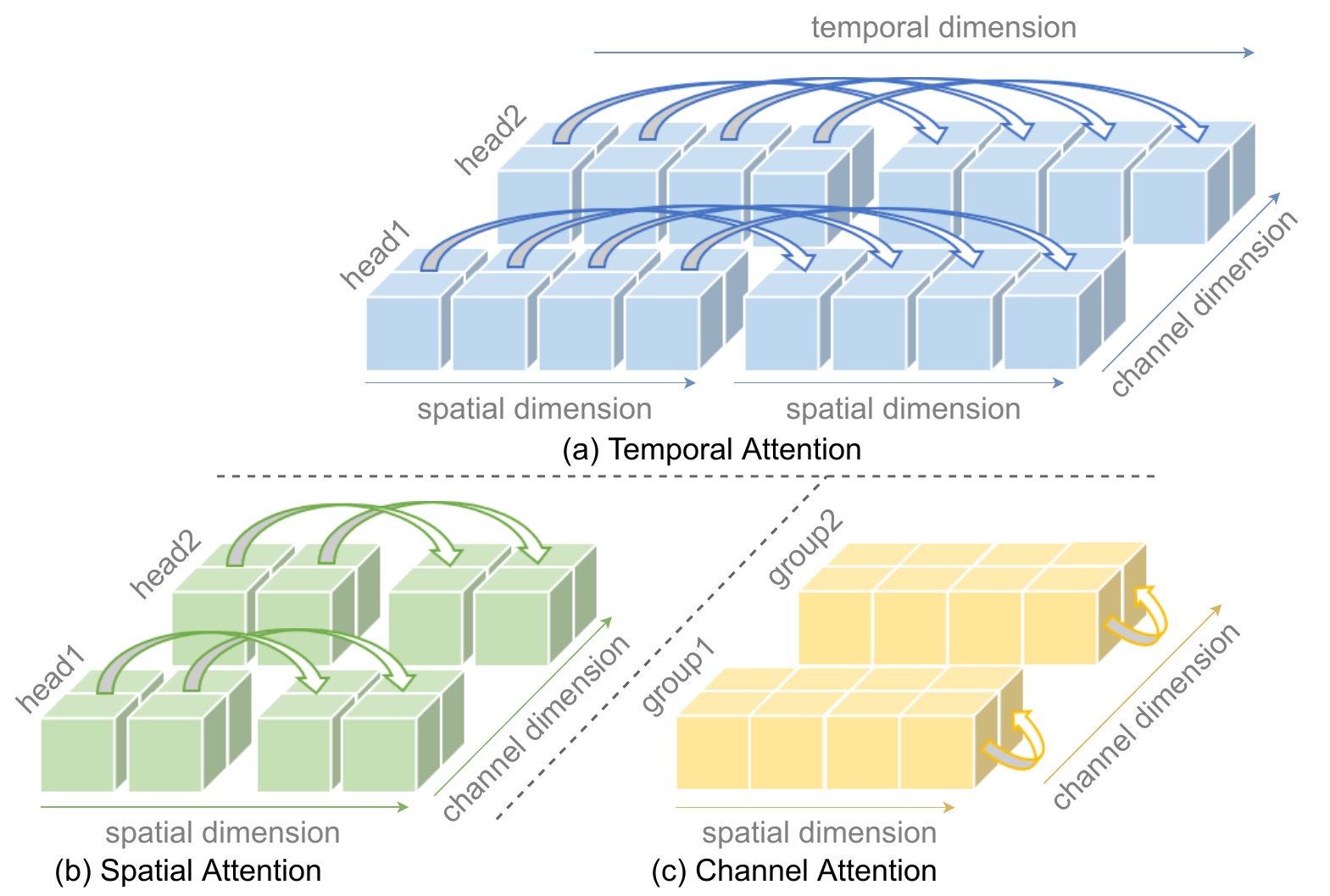}}
\end{center}
\vspace{-1.0em}
\caption{(a) Temporal attention allows cross-frame interaction tokens to enable long-term modeling. (b) Spatial attention partitions the spatial tokens into global grids and performs the unshuffle operation to implement global spatial interaction. (c) Channel attention is performed in each channel group with linear computational effort. In this work, we alternately use three types of attention to learn short- and long-range spatiotemporal information.}
\label{fig:triplet_dim}
\end{figure}

\begin{figure*}[t]
\begin{center}
\scalebox{1.0}{
\includegraphics[width=0.9\textwidth]{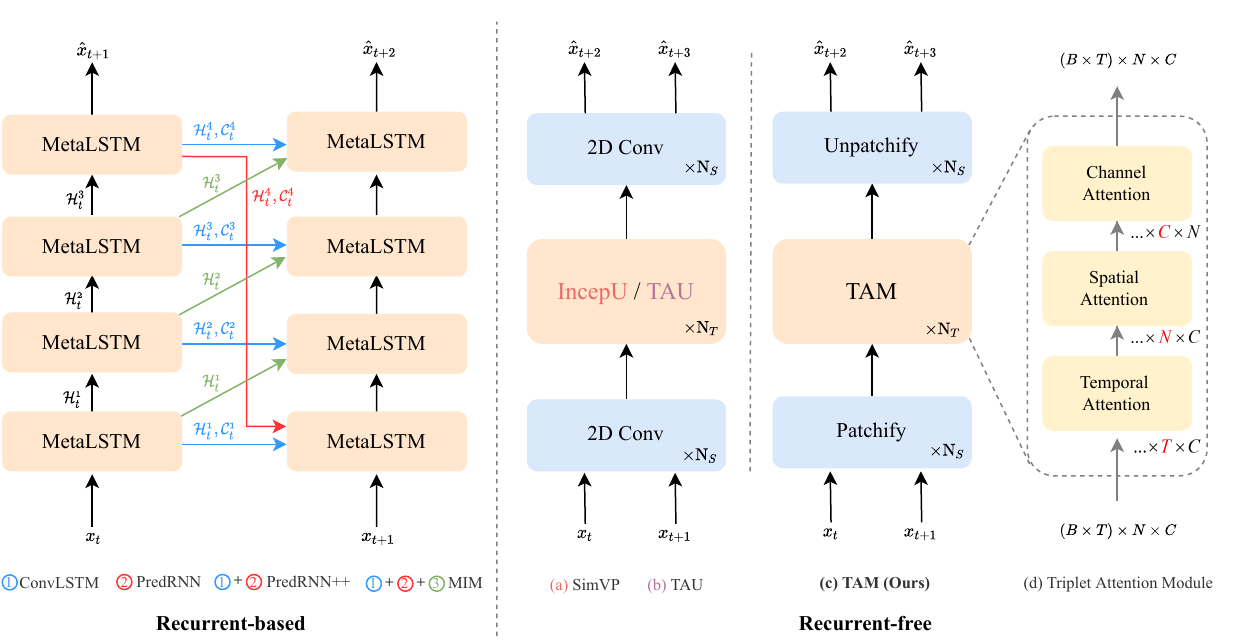}
}
\end{center}
\vspace{-2.0em}
\caption{Two typical spatiotemporal predictive learning frameworks. The recurrent-based methods extract spatiotemporal dependencies by cooperating with recurrent units MetaLSTM, and the information transmission between these units. The recurrent-free such as (a) SimVP and (b) TAU extract spatiotemporal features through Inception and temporal attention unit. In contrast, Our approach implements attention mechanisms in each dimension to learn more sophisticated spatiotemporal dependencies.}
\label{fig:triplet_arch}
\end{figure*}

Struggling with the inherent complexity and randomness of future events, spatiotemporal predictive learning has progressively evolved into two approaches, \textit{recurrent-based} and \textit{recurrent-free} frameworks shown in Figure~\ref{fig:triplet_arch}. The recurrent-based methods~\cite{chang2021mau,wang2022predrnn,wang2019memory} dominate the task due to their superior temporal modeling ability. Many mainstream models~\cite{wang2022predrnn,tang2023swinlstm} with stacked recurrent units capture the temporal dependencies. Inspired by the success of long short-term memory (LSTM) networks~\cite{hochreiter1997long} in sequential modeling, ConvLSTM~\cite{shi2015convolutional}, PredRNN~\cite{wang2017predrnn}, PredRNN++~\cite{wang2018predrnn++}, and MIM~\cite{wang2019memory} propose various LSTM variants, called MetaLSTM, such as ConvLSTM, ST-LSTM, Causal LSTM, and MIM-LSTM. Thus we abstract the general framework of recurrent-based models on the left side of Figure~\ref{fig:triplet_arch}, which consists of two main parts: (i) various LSTM variants MetaLSTM; (ii) mode of feature information transmission across different time steps. While the recurrent-based framework is superior in predictive performance, non-parallelizable and computational inefficiency limits its further application. Recently, recurrent-free methods~\cite{tan2022simvp,tan2023temporal,hu2023dynamic} with the parallelizable advantage have been proposed for spatiotemporal learning. As shown in Figure~\ref{fig:triplet_arch}(a)(b), we demonstrated the recurrent-free framework representing SimVP~\cite{gao2022simvp} and TAU~\cite{tan2023temporal}, which also consists of two main parts: (i) spatial encoder-decoder; (ii) latent feature spatiotemporal translator. Despite greater computational efficiency, the above methods still have performance gaps in some scenarios compared to the recurrent-based model due to the irrobust modeling of intra- and inter-frame variations.

In this work, we present an innovative triplet attention mechanism that is able to learn short- and long-range sophisticated spatiotemporal dependencies while maintaining computational efficiency. We implemented the Triplet Attention Module (TAM) with an elegant yet simple manner, applying self-attention to the \textit{permutation} of the token matrix, as shown in Figure~\ref{fig:triplet_dim}. TAM is decomposed into temporal, spatial, and channel-level
attention to capturing temporal and spatial evolution. Specifically, temporal attention modeling inter-frame dynamics, while spatial and channel attention modeling intra-frame static features. We improve spatiotemporal prediction learning by replacing the dominant recurrent unit with a parallelizable pure attention framework. Combining the advantages of a parallelizable framework with the power of Transformer, we implement our proposed TAM blocks and surprisingly find the derived model achieves state-of-the-art in an extensive multi-scenario prediction, including synthetic moving object trajectory prediction, traffic flow prediction, driving scene prediction, and human motion capture. We outline our key contributions as follows:
\begin{itemize}
\item We propose the novel \textit{Triplet Attention Transformer} for spatiotemporal predictive learning, which seamlessly integrates intra- and inter-frame feature interaction to obtain powerful representation ability. 

\item We propose a parallelizable Triplet Attention Module (TAM), which enables models to learn complex short-term and long-term spatiotemporal dependencies through alternating use of \textit{temporal}, \textit{spatial}, and \textit{channel-level} attention.

\item We conduct extensive experiments that outperform existing recurrent-based and recurrent-free networks, achieving state-of-the-art results on Kitti\&Caltech, Human3.6M, TaxiBJ, and Moving MNIST datasets.
\end{itemize}

\section{Related Work}
\paragraph{Self-Supervised Learning.} Despite the notable strides made with supervised learning methods on large labeled datasets, the limited labeled data constrains artificial intelligence development. In contrast, self-supervised learning, using plentiful unlabeled data, offers a promising route toward achieving human-level intelligence. Self-supervised learning creates guiding signals from the data itself via pretext tasks, enabling models to learn data representation. Early visual self-supervised tasks involved colorization~\cite{zhang2016colorful}, inpainting~\cite{pathak2016context}, rotation~\cite{gidaris2018unsupervised}, and jigsaw~\cite{noroozi2016unsupervised}. Contrastive learning~\cite{grill2020bootstrap,wang2020understanding,zbontar2021barlow}, while dominant, has limitations on small datasets due to its pair-making process. Masked reconstruction learning~\cite{devlin2019bert,lewis2020bart,he2022masked}, which predicts hidden parts from visible ones, is successful in natural language processing but challenging in visual tasks. In contrast to image-level methods, spatiotemporal predictive learning, a burgeoning self-supervised approach, emphasizes video-level information. It predicts upcoming frames by learning from previous ones, thus allowing the model to efficiently segregate foreground and background based on inherent motion dynamics.

\paragraph{Spatiotemporal Predictive Learning.} Recent strides in recurrent-based models have provided valuable insights into spatiotemporal predictive learning. ConvLSTM~\cite{shi2015convolutional}, a pioneering work, integrates convolutional networks into LSTM architecture. PredRNN~\cite{wang2017predrnn} proposes a spatio-temporal LSTM (ST-LSTM) based on vanilla ConvLSTM modules to model spatial and temporal variations. PredRNN++~\cite{wang2018predrnn++} proposes a Casual-LSTM to connect spatial and temporal memories and a gradient highway unit to mitigate the gradient vanishing. MIM~\cite{wang2019memory} using differential information between hidden states for better non-stationarity handling. E3D-LSTM~\cite{wang2018eidetic} incorporating 3D convolutions into LSTM architecture. PredRNNv2~\cite{wang2022predrnn} proposes a curriculum learning strategy and memory decoupling loss for enhanced performance. MAU~\cite{chang2021mau} designs a motion-aware unit to capture motion information. SwinLSTM~\cite{tang2023swinlstm} integrates the Swin Transformer~\cite{liu2021swin} module into the LSTM architecture for better spatiotemporal modeling. Recently, recurrent-free models have achieved superior performance with the advantage of parallelization. SimVP~\cite{gao2022simvp}, a seminal work, applies blocks of Inception modules with a UNet architecture to learn the temporal evolution. TAU~\cite{tan2023temporal} proposed the temporal attention unit on this basis to capture time evolution. DMVFN~\cite{hu2023dynamic} proposes a dynamic multi-scale voxel flow network to achieve better prediction performance. Although these recurrent-free methods have achieved great success, their performance is still inferior to recurrent-based in certain scenarios. To this end, we use pure parallelizable attention mechanisms to learn more sophisticated spatiotemporal dependencies.

\paragraph{Vision Transformer.} Vision
Transformer~\cite{dosovitskiy2020image} (ViT) demonstrates exceptional performance across various vision tasks. To enhance its efficiency and effectiveness in image classification tasks,  a series of ViT-based approaches have been proposed. Swin Transformer~\cite{liu2021swin} employs local attention windows and implements shift operations to augment window-based interactions. DaViT~\cite{ding2022davit} introduces a dual self-attention mechanism aimed at capturing global context with linear computational complexity. Due to the remarkable performance of ViT, researchers are now using them to understand video content. Uniformer~\cite{li2023uniformer} organically unifies convolution and self-attention to solve local redundancy and global dependency. TimeSformer~\cite{bertasius2021space} and ViViT~\cite{arnab2021vivit} explore separate strategies for temporal and spatial attention, achieving excellent outcomes. MViT~\cite{fan2021multiscale} introduces multiscale features for video sequences, and Video Swin Transformer~\cite{liu2022video} adapts the model to 3D settings. However, most existing models concentrate on video classification and works about video prediction using ViT are still limited. Is there a solution that combines the strengths of recurrent-based and recurrent-free architecture and takes advantage of the high performance of ViT? Therefore, we propose a triplet attention transformer for efficient spatiotemporal predictive learning.

\section{Preliminaries}
\subsection{Problem Definition}
Given $X^{t: T}_{in}=\{X_{t}, \ldots, X_{T}\}$, the objective is to predict the most reasonable sequences of length $T^{\prime}$ in the future, denoted as $X^{T+1: T+T^{\prime}}_{out}=\{\widehat{X}_{T+1}, \ldots, \widehat{X}_{T+T^{\prime}}\}$. We represent the spatiotemporal sequences as a four-dimensional tensor, $i.e.$, $X^{t: T}_{in} \in \mathbb{R}^{T \times C \times H \times W}$, where $C$, $T$, $H$, and $W$ denote channel, temporal or frames, height and width, respectively. The model with learnable parameters $\theta$ learns a mapping $\mathcal{F}_{\theta}: \mathcal{X}^{t: T}_{in} \mapsto \mathcal{X}^{T+1: T+T^{\prime}}_{out}$ by exploring spatiotemporal dependencies. Concretely, we use the stochastic gradient descent algorithm to learn the mapping $\mathcal{F}_{\theta}$ and find a set of parameters $\theta^{\star}$, which minimize the difference between the prediction and the ground-truth, the optimal parameters $\theta^{\star}$ are:
\begin{align}
\theta^{\star}=\arg \min _{\theta} \mathcal{L}\left(\mathcal{F}_{\theta}\left(X^{t: T}_{in}\right), X^{T+1: T+T^{\prime}}_{out}\right) ,
\end{align}
where $\mathcal{L}$ denote a loss function. In this paper, we adopt the vanilla Mean Squared Error (MSE) as our loss metric. 

% We focus on video prediction as our primary experimental domain, wherein the observed data consists of RGB images with three channels. In alternative domains, such as traffic flow prediction, the observed data manifests as tensors with single or multiple channels.

\begin{figure*}[ht]
\begin{center}
\scalebox{1.0}{
\includegraphics[width=0.98\textwidth]{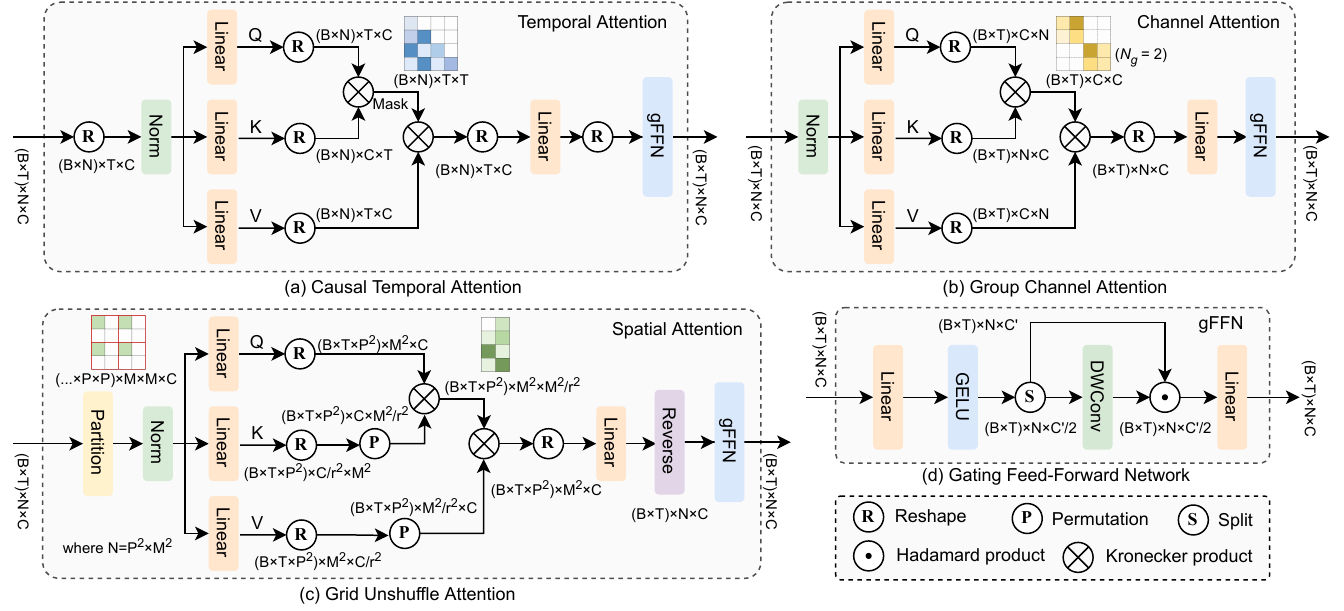}
}
\end{center}
\vspace{-1.0em}
\caption{The detailed architecture for triplet attention module. It contains three attention blocks: (a) Causal temporal attention, (b) Group channel attention, and (c) Grid unshuffle attention. By alternately using the three types of attention, our model enjoys the benefit of capturing both spatial dependency and temporal variation. (d) Gating feed-forward network reduces redundant information in channels.}
\label{fig:triplet_attn}
\end{figure*}

\subsection{Self-Attention Mechanism}
Assume a visual feature with dimension $\mathbb{R}^{N \times C}$, where $N$ is the number of total patches and $C$ is the number of total channels. Simply applying the standard global self-attention leads to quadratic complexity about input tokens. It is defined as:
\begin{equation}
\begin{aligned}
\mathcal{A}(\mathbf{Q}, \mathbf{K}, \mathbf{V}) & =\operatorname{Concat}\left(\text{head}_1, \ldots, \text{head}_{N_h}\right), \\
\text{where } \text{head}_i & =\operatorname{Attention}\left(\mathbf{Q}_i, \mathbf{K}_i, \mathbf{V}_i\right), \\
& =\operatorname{Softmax}\left[\frac{\mathbf{Q}_i\left(\mathbf{K}_i\right)^{\mathrm{T}}}{\sqrt{C_h}}\right] \mathbf{V}_i,
\end{aligned}
\end{equation}
where $\mathbf{Q}_i=X_i \mathbf{W}_i^Q, \mathbf{K}_i=X_i \mathbf{W}_i^K$, and $\mathbf{V}_i=X_i \mathbf{W}_i^V$ are $\mathbb{R}^{N \times C_h}$ dimensional visual tokens with $N_h$ heads, $X_i$ denotes the $i_{t h}$ head of the input tokens and $\mathbf{W}_i$ denotes the projection weights of the $i_{t h}$ head for $\mathbf{Q}, \mathbf{K}, \mathbf{V}$, and $C=C_h * N_h$. Please note that we omit the output projection $\mathbf{W}^O$. It's noteworthy that, due to potential large values of $N$ ($e.g.$, $64 \times 64$), the computational implications can be significant.

In this paper, we alternatively arrange causal temporal attention, grid unshuffle attention, and group channel attention to learn more sophisticated spatiotemporal dependencies with less complexity, as shown in Figure~\ref{fig:triplet_attn}.

\section{Proposed Method}
We approach the concept of self-attention from an alternative perspective, proposing a Triplet Attention Module (TAM) that integrates temporal, spatial, and channel-level attention for optimized spatiotemporal predictive learning. Striving for simplicity, the model follows the general framework in Figure~\ref{fig:triplet_arch}(c), incorporates the \textit{patchify} and \textit{unpatchify} module which comprises vanilla 2D convolutional and transposed convolutional layers. TAM leverages the self-attention mechanism across varying dimensions, as outlined in Figure~\ref{fig:triplet_dim}, we introduce three distinct attention modules: \textit{Causal Temporal Attention}, \textit{Grid Unshuffle Attention}, and \textit{Group Channel Attention}. In the middle layer, the repeated stacking of TAM facilitates the learning of both short-term and long-term complex spatiotemporal dependencies. Therefore, a comprehensive discourse on these modules is provided subsequently. 

\subsection{Causal Temporal Attention}
Previous vision-based self-attention~\cite{zhu2023biformer,tu2022maxvit,liu2021swin}, tokens have been defined using pixels or patches, emphasizing spatial dimensions. Instead of spatial attention, we apply attention mechanisms on the temporal-level tokens to capture long-term dependencies. This allows temporal tokens to interact with inter-frame information more efficiently. Although our approach employs a non-autoregressive framework, it can be easily extended to parallelizable autoregression compared to non-parallelizable recurrent-based models. Specifically, we achieve this by masking out (setting to $-\infty$) all values of the upper triangle of the attention matrix to prevent previous frames from seeing subsequent frames, as shown in Figure~\ref{fig:triplet_attn}(a).

Simple permutation of feature dimensions can obtain vanilla temporal-level attention. Formally, let $T$ denote the number of frames, $N$ the number of patches, and $C$ the number of channels. Therefore, the token of each frame is designed to interact across other frames. It is defined as: 
\begin{equation}
\begin{aligned}
\mathcal{A}_{\textit{temporal}}(\mathbf{Q}, \mathbf{K}, \mathbf{V})=\left\{\mathcal{A}\left(\mathbf{Q}_i, \mathbf{K}_i, \mathbf{V}_i\right)\right\}_{i=0}^{N}, \\
\mathcal{A}\left(\mathbf{Q}_i, \mathbf{K}_i, \mathbf{V}_i\right)  =\operatorname{Softmax}\left[\frac{\mathcal{M}(\mathbf{Q}_i^{\mathrm{T}} \mathbf{K}_i)}{\sqrt{C_k}}\right] \mathbf{V}_i^T,
\end{aligned}
\end{equation}
where $\mathbf{Q}_i, \mathbf{K}_i, \mathbf{V}_i \in \mathbb{R}^{T \times C}$ are frame-wise image-level queries, keys, and values, and $\mathcal{M}$ represents the mask operation. The aforementioned equation can be adapted to a multi-head version by dividing the channels into several groups. More details are shown in Figure~\ref{fig:triplet_attn}(a). 

% \paragraph{Gating Feed-Forward Network.} The conventional feed-forward network (FFN) uses two linear projection layers and non-linear activation for feature extraction but overlooks spatial information and often contains redundant channel information. Addressing this, we introduce the gating feed-forward network (gFFN) (Figure~\ref{fig:triplet_attn}(d)). This network integrates a simple gating mechanism, employing depth-wise convolution and the Hadamard product. The feature map is divided along the channel dimension for specialized processing. For an input $X \in \mathbb{R}^{N \times C}$, the gFFN operates as:
% \begin{equation}
% \begin{aligned}
% & X^{\prime}=\sigma\left(\mathbf{W}_p^1 X\right), \quad\left[X_1^{\prime}, X_2^{\prime}\right]=X^{\prime} \\
% & \operatorname{gFFN}(X)=\mathbf{W}_p^2\left(X_1^{\prime} \odot\left(\mathbf{W}_d X_2^{\prime}\right)\right)
% \end{aligned}
% \end{equation}
% where using linear projections $\mathbf{W}_p^1$ and $\mathbf{W}_p^2$, the GELU function $\sigma$, and depth-wise convolution parameters $\mathbf{W}_d$, both $X_1^{\prime}$ and $X_2^{\prime}$ occupy $\mathbb{R}^{H \times W \times \frac{C^{\prime}}{2}}$ space. Here, $C^{\prime}$ is the hidden dimension in gFFN.

\subsection{Grid Unshuffle Attention}
\begin{figure}[t]
\begin{center}
\scalebox{0.4}{\includegraphics[width=0.98\textwidth]{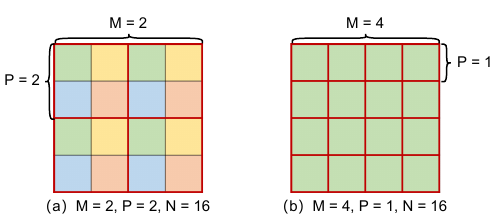}}
\end{center}
\vspace{-1.5em}
\caption{Compressing channels permutate spatial tokens into channel tokens allowing larger grid size $M$ with less computation.}
\label{fig:grid_size}
\end{figure}

To address the quadratic complexity with the number of input tokens in ViT, we adopted the approach from prior research~\cite{mehta2021mobilevit,tu2022maxvit} involving gridded feature maps to aggregate global tokens. A smaller grid size $M$ will result in a gridding effect, Figure~\ref{fig:grid_size} illustrates our use of the unshuffle operation to permutate the spatial token to the channel token for an expanded grid size $M$. Specifically, as depicted in Figure~\ref{fig:triplet_attn}(c),  for an input feature map $X \in \mathbb{R}^{N \times C}$ and an unshuffle factor $r$, we partition $X$ into $M \times M$ non-overlapping windows. Subsequently, we gather tokens at identical locations in each window, denoted as $X_p \in \mathbb{R}^{P^2 \times M^2 \times C}$ where $P \times P$ represents the total tokens per window and $N = P^2 \times M^2$. We then employ three linear layers, $\mathbf{W}_i^Q$, $\mathbf{W}_i^K$, and $\mathbf{W}_i^V$, to obtain $\mathbf{Q}_i$, $\mathbf{K}_i$, and $\mathbf{V}_i$:
\begin{equation}
\mathbf{Q}_i, \mathbf{K}_i, \mathbf{V}_i=\mathbf{W}_i^Q(X_p), \mathbf{W}_i^K(X_p), \mathbf{W}_i^V(X_p),
\end{equation}
Here, $\mathbf{Q}_i$ maintains its channel dimension, whereas $\mathbf{W}_i^K$ and $\mathbf{W}_i^V$ reduce it to $C / r^2$, resulting in $\mathbf{K}_i \in \mathbb{R}^{P^2 M^2 \times C / r^2}$ and $\mathbf{V}_i \in \mathbb{R}^{P^2 M^2 \times C / r^2}$. Spatial tokens in these matrices are permuted to channel tokens, producing $\mathbf{K}_i^p \in$ $\mathbb{R}^{C \times M^2 / r^2}$ and $\mathbf{V}_i^p \in \mathbb{R}^{M^2 / r^2 \times C}$. Using $\mathbf{Q}_i$ with these permuted tokens, we execute the self-attention operation, enabling larger grid sizes (e.g., $24 \times 24$) with fewer computations than $8 \times 8$, yet achieving superior performance. It is defined as:
\begin{equation}
\begin{aligned}
\mathcal{A}_{\textit{spatial}}(\mathbf{Q}, \mathbf{K}, \mathbf{V})=\left\{\mathcal{A}\left(\mathbf{Q}_i, \mathbf{K}_i^p, \mathbf{V}_i^p\right)\right\}_{i=0}^P, \\
\mathcal{A}\left(\mathbf{Q}_i, \mathbf{K}_i^p, \mathbf{V}_i^p\right) =\operatorname{Softmax}\left[\frac{(\mathbf{Q}_i (\mathbf{K}_i^p)^{\mathrm{T}}}{\sqrt{C_k}} + \mathbf{B}\right] \mathbf{V}_i^p,
\end{aligned}
\end{equation}
where $\mathbf{B}$ is the relative position embedding.

\subsection{Group Channel Attention}
Channel Attention is similar to the above modules, as illustrated in Figure~\ref{fig:triplet_dim}(c), we employ self-attention in the channel dimension. Notably, in many scenarios, the number of channels will be higher ($e.g.$, $C = 256$). To mitigate the inherent quadratic complexity of self-attention concerning the channel dimension, we group channels into multiple groups and perform self-attention within each group. Formally, let $N_g$ denote the number of groups and $C_g$ the number of channels in each group, we have $C=N_g * C_g$. More details are shown in Figure~\ref{fig:triplet_attn}(b). It is defined as:
\begin{equation}
\begin{aligned}
\mathcal{A}_{\textit{channel}}(\mathbf{Q}, \mathbf{K}, \mathbf{V}) & =\left\{\mathcal{A}\left(\mathbf{Q}_i, \mathbf{K}_i, \mathbf{V}_i\right)^T\right\}_{i=0}^{N_g}, \\
\mathcal{A}\left(\mathbf{Q}_i, \mathbf{K}_i, \mathbf{V}_i\right) & =\operatorname{Softmax}\left[\frac{\mathbf{Q}_i^{\mathrm{T}} \mathbf{K}_i}{\sqrt{C_g}}\right] \mathbf{V}_i^T.
\end{aligned}
\end{equation}
where $\mathbf{Q}_i, \mathbf{K}_i, \mathbf{V}_i \in \mathbb{R}^{N \times C_g}$ as channel-grouped queries, keys, and values. To accommodate frames with different sizes, the projection layers $\mathbf{W}$ remain performed along the channel dimension. We also use conditional positional encoding~\cite{chu2021conditional} (CPE) to provide location information.
%-------------------------------------------------------------------------

\begin{figure*}[h]
\begin{center}
\scalebox{1.0}{
\includegraphics[width=0.98\textwidth]{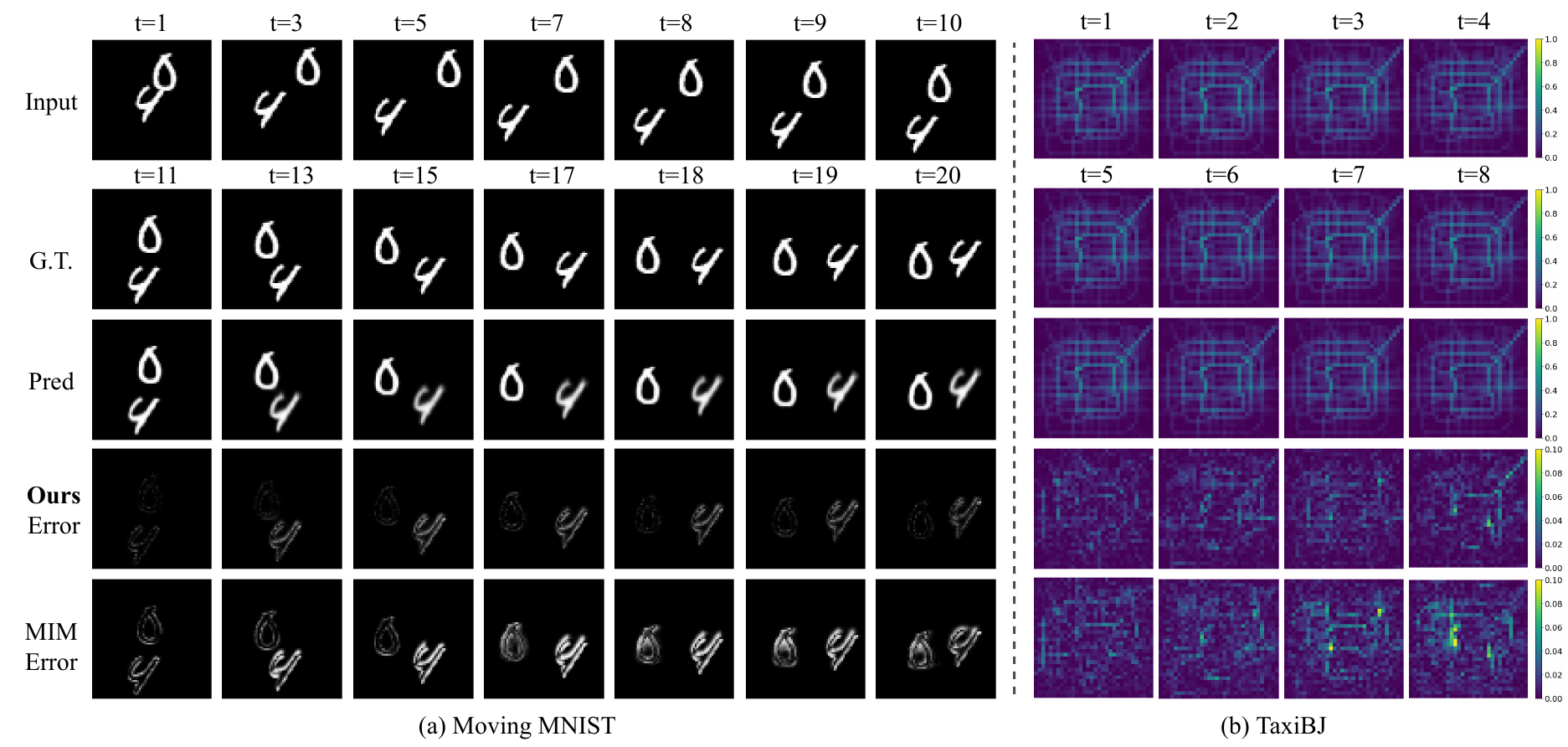}
}
\end{center}
\vspace{-1.5em}
\caption{Qualitative visualizations on (a) Moving MNIST and (b) TaxiBJ, where prediction error = $|$ground truth - prediction$|$.}
\label{fig:mmnist_taxibj}
\end{figure*}

\section{Experiments}
\label{sec:setting}
\paragraph{Multi-Scenario Examination.} Our model is quantitatively evaluated across expansive real-world scenarios with diverse scales, including traffic flow prediction, driving scene prediction, and human motion capture. For synthetic data scenarios such as Moving MNIST~\cite{srivastava2015unsupervised}, we also offer comprehensive experiments. Each dataset gathers from various domains, from micro to macro scales. The details of dataset statistics are shown in Table~\ref{tab:dataset}. For more experiments, please refer to the Supplementary Materials.

\begin{itemize}
\item \textbf{Synthetic Moving Object Trajectory Prediction.} The \textit{Moving MNIST} dataset~\cite{srivastava2015unsupervised} as a foundational benchmark that has been widely employed in various studies. This dataset is comprised of video sequences in which two digits traverse across a frame of dimensions $64 \times 64$ pixels. For each digit, its velocity is determined by two factors: (i) a direction randomly chosen from a unit circle; (ii) a magnitude arbitrarily selected from a predefined range.
\item \textbf{Traffic Flow Prediction.} Efficient traffic governance and public safety depend on accurate crowd dynamics prediction. We employ the \textit{TaxiBJ} dataset~\cite{zhang2017deep}, involving taxi GPS trajectories from Beijing bifurcated into two channels: inflow and outflow. This dataset has a 30-minute interval and a spatial granularity of $32 \times 32$. Our approaches to data preprocessing, model training, and performance assessment are congruent with protocols by PredNet~\cite{guen2020disentangling} and MIM~\cite{wang2019memory}.
\item \textbf{Driving Scene Prediction.} In autonomous driving, predicting future dynamics is critically important in complex and non-stationary environments. We employ two datasets for evaluation purposes: \textit{KITTI}~\cite{geiger2013vision} extensively utilized in the fields of autonomous driving and robotics; \textit{Caltech Pedestrian}~\cite{dollar2009pedestrian} specializes in pedestrian detection. We train our model on the \textit{KITTI} dataset and evaluate performance on the \textit{Caltech Pedestrian} benchmark.
\item \textbf{Human Motion Capture.} Predicting human motion remains a formidable challenge due to the considerable variability across individual behaviors and actions. In our study, we utilize the \textit{Human3.6M}~\cite{ionescu2013human3} dataset, encompassing high-resolution motion capture videos. Following previous work settings~\cite{guen2020disentangling}, we employ four observed frames to predict the subsequent four frames.
\end{itemize}

% \item \textbf{Meteorological Forecasting.} 
% Predicting global weather patterns is pivotal for advancing spatiotemporal predictive learning. In this study, we employ the WeatherBench~\cite{rasp2020weatherbench} dataset, an extensive resource for weather forecasting spanning from 1979 to 2018. This dataset is re-gridded to resolutions of $5.625^{\circ}$ (corresponding to $32 \times 64$ grid points) and $1.40625^{\circ}$ (corresponding to $128 \times 256$ grid points). We segment the dataset to train the model using data from 2010 to 2015, validate it with data from 2016, and test it using data from 2017 to 2018. Forecasts are made at one-hour intervals, leveraging climatic parameters from the preceding 12 hours---namely, temperature, humidity, wind component, and cloud cover---to predict conditions for the upcoming 12 hours.

\paragraph{Evaluation Metrics.} We evaluate the performance of the proposed model using various metrics. For pixel-wise error, we consider mean squared error (MSE), mean absolute error (MAE), and root mean squared error (RMSE). Structural similarity index measure (SSIM) and peak signal-to-noise ratio (PSNR) are used for similarity evaluation. 

\begin{table}[h]
\centering
\small
\setlength{\tabcolsep}{2.0mm}
\resizebox{\linewidth}{!}{
\begin{tabular}{cccccccc}
\toprule
Dataset & Train & Test & $C$ & $H$ & $W$ & $T$ & $T^{\prime}$ \\
\midrule
Kitti\&Caltech & 3,160    & 3,095   & 3 & 128 & 160  & 10  & 1 \\
Human3.6M    &  73,404 & 8,582 & 3 & 256  & 256   & 4  & 4 \\
TaxiBJ       &  20,461 & 500  & 2 & 32  & 32   & 4   & 4 \\
% WeatherBench & 2,167 & 706 & 1 & 32  & 64   & 12  & 12 \\ 
Moving MNIST &  10,000 & 10,000  & 1 & 64  & 64   & 10   & 10 \\
\bottomrule
\end{tabular}}
\caption{The details of dataset statistics. We detail the number of samples, the input frames denoted as $T$, and the predicted frames represented as $T^{\prime}$ for both the training and testing subsets.}
\vspace{-1.0em}
\label{tab:dataset}
\end{table}

\paragraph{Implementation Details.} We use the PyTorch framework on a single NVIDIA-V100 GPU for our proposed method. The model trains with mini-batches of 16 video sequences using the AdamW optimizer, the OneCycle learning rate scheduler, and a weight decay of $5 e^{-2}$. The learning rate, selected from $\{1 e^{-2}, 5 e^{-3}, 1 e^{-3}\}$, ensures stable training. We employ stochastic depth as regularization.

\begin{figure*}[t]
\begin{center}
\scalebox{1.0}{
\includegraphics[width=0.98\textwidth]{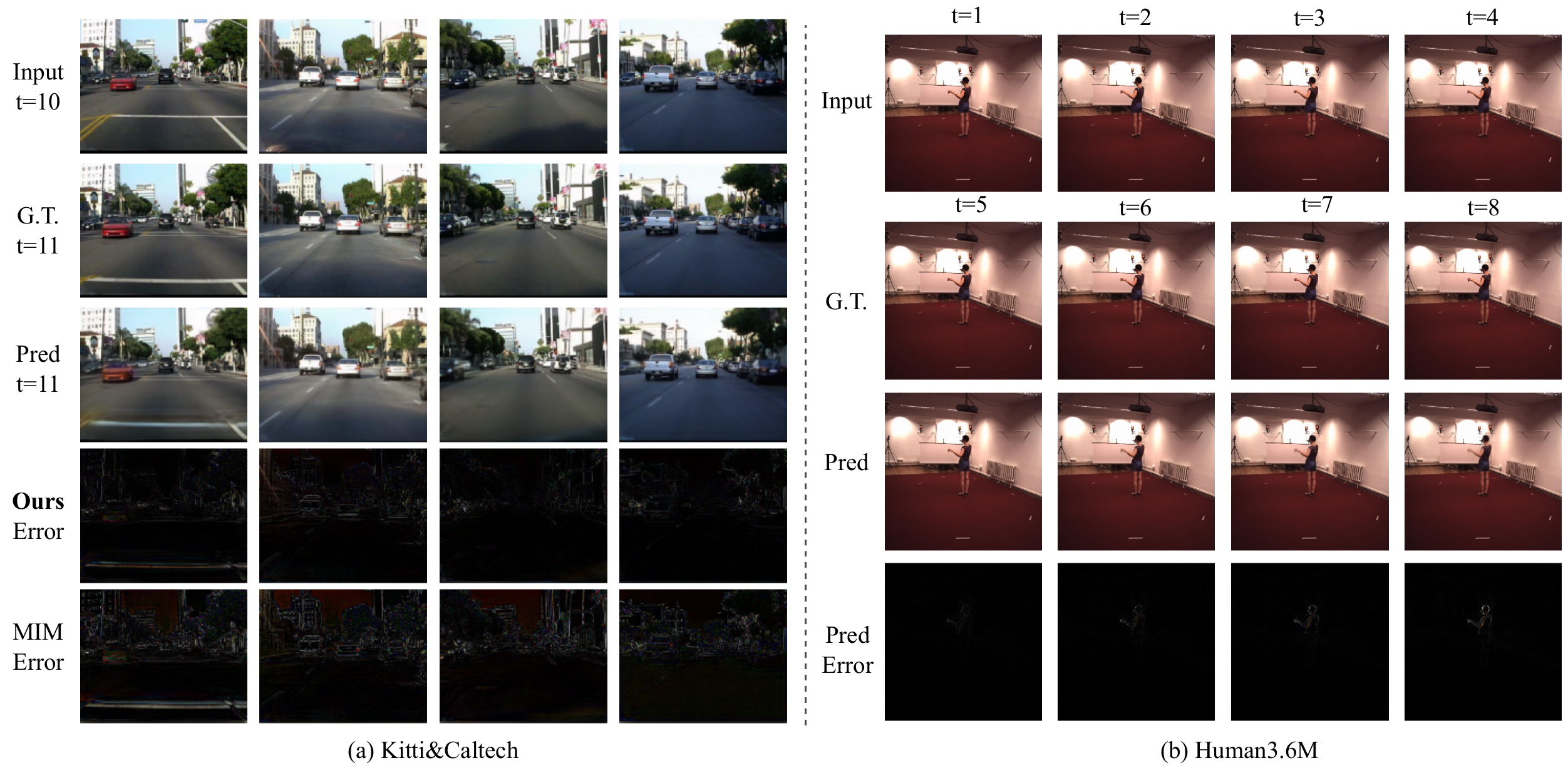}
}
\end{center}
\vspace{-1.5em}
\caption{Qualitative visualizations on (a) Kitti\&Caltech and (b) Human3.6M, where prediction error = $|$ground truth - prediction$|$.}
\label{fig:kitti_human}
\end{figure*}

\subsection{Synthetic Moving Object Trajectory Prediction}
\paragraph{Moving MNIST.} This dataset is a standard benchmark for evaluating spatiotemporal predictive learning methods. We compare our proposed approach with various recent strong baselines. The quantitative results are detailed in Table~\ref{tab:mmnist}, and visualizations of the predictions can be found in Figure~\ref{fig:mmnist_taxibj}(a). Notably, our method exceeds all baselines under four separate metrics. Compared to ConvLSTM~\cite{shi2015convolutional}, our method reduces the MSE from 103.3 to 17.55 and increases the SSIM from 0.707 to 0.966. In contrast to MIM~\cite{wang2019memory}, our method can accurately predict motion trajectories and appearances of two digits. We also tried to experiment with autoregressive (w/ AR) methods of the recurrent-based architectures, finding that while it improved prediction quality, more time was needed to complete the training.

\begin{table}[htbp]
\centering
\setlength{\tabcolsep}{2.0mm}
\resizebox{\linewidth}{!}{
\begin{tabular}{cccccc}
\toprule
\multicolumn{6}{c}{Moving MNIST ($10 \rightarrow 10$ frames)} \\
Method & Reference & MSE & MAE & SSIM & PSNR \\ 
& & $\downarrow$ & $\downarrow$ & $\uparrow$ & $\uparrow$ \\
\midrule
ConvLSTM~\cite{shi2015convolutional} & NIPS'2015 & 103.3 & 182.9 & 0.707 & 16.17\\
PredRNN~\cite{wang2017predrnn} & NIPS'2017 & 56.8 & 126.1 & 0.867 & 19.12\\
PredRNN++~\cite{wang2018predrnn++} & ICML'2018 & 46.5 & 106.8 & 0.898 & 20.11\\
MIM~\cite{wang2019memory} & CVPR'2019 & 44.2  & 101.1 & 0.910 & 20.31\\
E3D-LSTM~\cite{wang2018eidetic} & ICLR'2019 & 41.3  & 87.2 & 0.910 & 20.70\\
MAU~\cite{chang2021mau} & NIPS'2021 & 27.6  & 86.5 & 0.937 & 22.59 \\
PredRNNv2~\cite{wang2022predrnn} & TPAMI'2022 & 48.4  & 129.8 & 0.891 & 20.12 \\
SimVP~\cite{gao2022simvp} & CVPR'2022 & 23.8  & 68.9 & 0.948 & 23.19\\ 
\underline{TAU}~\cite{tan2023temporal} & CVPR'2023 & 19.8  & 60.3 & 0.957 & 24.53\\
DMVFN~\cite{hu2023dynamic} & CVPR'2023 & 123.6 & 179.9 & 0.814 & 16.15 \\
\hline
Ours & - & 17.55 & 59.81 & 0.960 & 25.08 \\ 
\textbf{Ours w/ AR} & - & \textbf{15.68}  & \textbf{51.85} & \textbf{0.966} & \textbf{25.71} \\ 
\bottomrule
\end{tabular}}
\caption{Quantitative results on the Moving MNIST dataset.}
\vspace{-1.0em}
\label{tab:mmnist}
\end{table}

\subsection{Traffic Flow Prediction}
\paragraph{Taxibj.} Traffic flow prediction presents significant challenges due to the unpredictability introduced by human behavior. Our method is evaluated using the TaxiBJ dataset~\cite{zhang2017deep}, which embodies the complex nature inherent in real-world traffic systems. The complexity of road networks and nonlinear temporal behaviors limit the efficacy of traditional forecasting methods.

Table~\ref{tab:taxibj} reports the quantitative results, while Figure~\ref{fig:mmnist_taxibj}(b) offers qualitative visualizations. To optimize the visual interpretation, the error scale is limited to 0.1 and focuses solely on the inflow case. Despite minor deviations between observed and future data frames, our model consistently yields precise forecasts compared to recurrent-based methods. Owing to the robust spatiotemporal relationships captured by the triplet attention module, our methodology sets new benchmarks across all evaluation metrics, suggesting its suitability for application in traffic flow prediction.

\begin{table}[htbp]
\centering
\small
\setlength{\tabcolsep}{2.0mm}
\resizebox{\linewidth}{!}{
\begin{tabular}{cccccc}
\toprule
\multicolumn{6}{c}{TaxiBJ ($4 \rightarrow 4$ frames)} \\
Method & Reference & MSE & MAE & SSIM & PSNR\\ 
& & $\times$ 100$\downarrow$ & $\downarrow$ & $\uparrow$ & $\uparrow$ \\
\midrule
ConvLSTM~\cite{shi2015convolutional} & NIPS'2015 & 48.5  & 17.7 & 0.978 & 37.38 \\
PredRNN~\cite{wang2017predrnn} & NIPS'2017 & 46.4  & 17.1 & 0.971 & 38.52 \\
PredRNN++~\cite{wang2018predrnn++} & ICML'2018 & 44.8  & 16.9 & 0.977 & 38.71 \\
MIM~\cite{wang2019memory} & CVPR'2019 & 42.9  & 16.6 & 0.971 & 38.71 \\
E3D-LSTM~\cite{wang2018eidetic} & ICLR'2019 & 43.2  & 16.9 & 0.979 & 38.75 \\
PhyDNet~\cite{guen2020disentangling} & CVPR'2020 & 41.9  & 16.2 & 0.982 & 39.18 \\ 
SimVP~\cite{gao2022simvp} & CVPR'2022 & 41.4  & 16.2 & 0.982 & 39.17 \\ 
PredRNNv2~\cite{wang2022predrnn} & TPAMI'2022 & 38.3  & 15.6 & 0.983 & 39.38 \\
\underline{TAU}~\cite{tan2023temporal} & CVPR'2023 & 34.4  & 15.6 & 0.983 & 39.50 \\ 
SwinLSTM~\cite{tang2023swinlstm} & ICCV'2023 & 43.1  & 17.3 & 0.977 & 38.71 \\ 
\hline
\textbf{Ours} & - & \textbf{31.3}  & \textbf{15.1} & \textbf{0.984} & \textbf{39.67} \\ 
\bottomrule
\end{tabular}}
\caption{Quantitative results in the TaxiBJ dataset.}
\vspace{-1.0em}
\label{tab:taxibj}
\end{table}

\subsection{Driving Scene Prediction}
\paragraph{Kitti\&Caltech.} The ability to generalize is important in artificial intelligence. Traditional supervised learning often has limitations when applied to diverse domains. In contrast, self-supervised learning methods, such as contrastive learning and masked reconstruction learning, aim to learn robust representations from unlabeled data. These models then evaluate generalization ability through downstream tasks. In this paper, we evaluated this ability across different datasets, where we train our model on the KITTI~\cite{geiger2013vision} and then evaluate its performance on the Caltech Pedestrian~\cite{dollar2009pedestrian}.

Figure~\ref{fig:kitti_human} presents our qualitative visualizations, while Table~\ref{tab:kitti_caltech} offers the quantitative results. Remarkably, our method not only surpasses all recurrent-based approaches but also establishes new state-of-the-art results. It can be seen from the prediction errors in the last two rows of Figure~\ref{fig:kitti_human}(a), that our model effectively predicts both lane lines and distant vehicles. Given its consistent accuracy in dealing with variations in lighting and lane lines, our approach shows promise for application in autonomous vehicles.

\begin{table}[htbp]
\centering
\small
\setlength{\tabcolsep}{2.0mm}
\resizebox{\linewidth}{!}{
\begin{tabular}{cccccc}
\toprule
\multicolumn{6}{c}{Kitti\&Caltech ($10 \rightarrow 1$ frames)} \\
Method & Reference & MSE & MAE & SSIM & PSNR\\ 
& & $\downarrow$ & $\downarrow$ & $\uparrow$ & $\uparrow$\\
\midrule
ConvLSTM~\cite{shi2015convolutional} & NIPS'2015 & 139.6  & 1583.3 & 0.9345 & 27.46 \\
PredRNN~\cite{wang2017predrnn} & NIPS'2017 & 130.4  & 1525.5 & 0.9374 & 27.81 \\
PredRNN++~\cite{wang2018predrnn++} & ICML'2018 & 129.6 & 1507.7 & 0.9453 & 27.89 \\
\underline{MIM}~\cite{wang2019memory} & CVPR'2019 & 127.4  & 1476.5 & 0.9461 & 27.98 \\
E3D-LSTM~\cite{wang2018eidetic} & ICLR'2019 & 200.6  & 1946.2 & 0.9047 & 25.45 \\
PhyDNet~\cite{guen2020disentangling} & CVPR'2020  & 312.2  & 2754.8 & 0.8615 & 23.26 \\
MAU~\cite{chang2021mau} & NIPS'2021 & 177.8  & 1800.4 & 0.9176 & 26.14 \\
SimVP~\cite{gao2022simvp} & CVPR'2022 & 160.2 & 1690.8 & 0.9338 & 26.81 \\ 
PredRNNv2~\cite{wang2022predrnn} & TPAMI'2022 & 147.8  & 1610.5 & 0.9330 & 27.12 \\
TAU~\cite{tan2023temporal} & CVPR'2023 & 131.1  & 1507.8 & 0.9456 & 27.83 \\
DMVFN~\cite{hu2023dynamic} & CVPR'2023 & 183.9  & 1531.1 & 0.9314 & 26.78 \\
\hline
\textbf{Ours} & - & \textbf{122.9} & \textbf{1416.2} & \textbf{0.9469} & \textbf{28.18} \\ 
\bottomrule
\end{tabular}}
\caption{Quantitative results in Kitti\&Caltech dataset.}
\vspace{-1.0em}
\label{tab:kitti_caltech}
\end{table}

\subsection{Human Motion Capture}
\paragraph{Human3.6M.} Predicting human motion is challenging due to both the need for high-resolution forecasting and the complexity introduced by human unpredictability. To provide a comprehensive evaluation from multiple perspectives, we employ MSE, MAE, SSIM, and PSNR as metrics. Table~\ref{tab:human} provides qualitative results, and it can be seen that our method consistently outperforms the recurrent-based methods and establishes a strong baseline. We also present the visualization in Figure~\ref{fig:kitti_human}(b), where the smaller prediction error reveals that our method can handle real-world dynamic scenarios.

\begin{table}[htbp]
\centering
\small
\setlength{\tabcolsep}{2.0mm}
\resizebox{\linewidth}{!}{
\begin{tabular}{cccccc}
\toprule
\multicolumn{6}{c}{Human3.6M ($4 \rightarrow 4$ frames)} \\
Method & Reference & MSE & MAE & SSIM & PSNR \\ 
& & $\downarrow$ & $\downarrow$ & $\uparrow$ & $\uparrow$ \\
\midrule
ConvLSTM~\cite{shi2015convolutional} & NIPS'2015 & 125.5  & 1566.7 & 0.9813 & 33.40 \\
PredRNN~\cite{wang2017predrnn} & NIPS'2017 & 113.2  & 1458.3 & 0.9831 & 33.94 \\
\underline{PredRNN++}~\cite{wang2018predrnn++} & ICML'2018 & 111.3 & 1454.4 & 0.9832 & 33.92 \\
MIM~\cite{wang2019memory} & CVPR'2019 & 112.1  & 1467.1 & 0.9829 & 33.97 \\
E3D-LSTM~\cite{wang2018eidetic} & ICLR'2019 & 143.3  & 1442.5 & 0.9803 & 32.52 \\
PhyDNet~\cite{guen2020disentangling} & CVPR'2020 & 125.7  & 1614.7 & 0.9804 & 33.05 \\
MAU~\cite{chang2021mau} & NIPS'2021 & 127.3  & 1577.0 & 0.9812 & 33.33 \\
SimVP~\cite{gao2022simvp} & CVPR'2022 & 115.8 & 1511.5 & 0.9822 & 33.73 \\ 
PredRNNv2~\cite{wang2022predrnn} & TPAMI'2022 & 114.9  & 1484.7 & 0.9827 & 33.84 \\
TAU~\cite{tan2023temporal} & CVPR'2023 & 113.3  & 1390.7 & \textbf{0.9839} & 34.03 \\ 
\hline
\textbf{Ours} & - & \textbf{108.4} & \textbf{1389.1} & \textbf{0.9839} & \textbf{34.18} \\ 
\bottomrule
\end{tabular}}
\caption{Quantitative results in Human3.6M dataset.}
\vspace{-1.0em}
\label{tab:human}
\end{table}

\subsection{Ablation Study}
\paragraph{Triplet attention layout.} We tested four configurations for our triplet attention mechanism: (i) temporal attention first; (ii) spatial attention first; (iii) channel attention first; and (iv) triplet attention parallel. Table~\ref{tab:ablation} results show similar performance across all settings, with a slight edge for `temporal attention first'.
\paragraph{Effects of different attention.} We evaluate the contribution of different attentions by removing certain attention to compare performance on the Human3.6M~\cite{ionescu2013human3} dataset. Table~\ref{tab:ablation} shows the results, it can be seen that temporal attention is relatively important as it models inter-frame dynamics, and the other two attentions model intra-frame static.  

\begin{table}[htbp]
\centering
\small
\setlength{\tabcolsep}{3mm}
\resizebox{\linewidth}{!}{
\begin{tabular}{lcc}
\toprule
Method & SSIM$\uparrow$ & PSNR$\uparrow$ \\ 
\midrule
Temporal Attention First & \textbf{0.9839} & \textbf{34.18} \\
Spatial Attention First & 0.9826 & 34.10 \\
Channel Attention First & 0.9824 & 34.07 \\
Triplet Attention Parallel & 0.9804 & 33.12 \\
\midrule
Triplet Attention Module & 0.9839 & 34.18 \\
- Temporal Attention & 0.9794 \textbf{(-0.0045)} & 32.77 \textbf{(-1.41)} \\
- Spatial Attention & 0.9809 (-0.0030) & 33.26 (-0.92) \\
- Channel Attention & 0.9813 (-0.0026) & 33.55 (-0.63) \\
\bottomrule
\end{tabular}}
\caption{Ablation study in Human3.6M dataset.}
\vspace{-1.0em}
\label{tab:ablation}
\end{table}

\section{Conclusion}
This work introduces a novel triplet attention mechanism comprising causal temporal attention, grid unshuffle attention, and group channel attention. This mechanism effectively learns short and long-range spatiotemporal dependencies while maintaining computational parallelism. The three self-attentions are complementary: (i) temporal attention captures temporal dependence due to the abstract representations in each temporal token; (ii) spatial and channel attention together refine intra-frame representation via fine-grained interactions across spatial and channel dimensions. Extensive validation across multiple scenarios demonstrates the superior performance of our method. In sum, our approach highlights the importance of both intra-frame and inter-frame variations and provides a novel perspective on spatiotemporal predictive learning.

% This paper presents an innovative triplet attention mechanism, containing causal temporal attention, grid unshuffle attention, and group channel attention, to learn short- and long-range sophisticated spatiotemporal dependencies while maintaining computational efficiency. We show that these three self-attentions complement each other: (i) since each temporal token contains abstract representations of inter-frame, the temporal attention naturally captures temporal dependence; (ii) spatial and channel attention combine to refine the intra-frame representation by performing fine-grained interactions across spatial and channel dimension. With extensive multi-scenario validation, our approach consistently demonstrates state-of-the-art performance. Overall, our approach emphasizes the significance of both intra-frame and inter-frame variations, enabling the model to capture long-term relationships and offering a novel, efficient perspective on spatiotemporal predictive learning.

\section*{Acknowledgements}
This work is supported by the CCF-Lenovo Bule Ocean Research Fund.

%%%%%%%%% REFERENCES
{\small
\bibliographystyle{ieee_fullname}
\bibliography{egbib}
}

\end{document}